\documentclass{article}

\usepackage[preprint]{neurips_2020}
\usepackage{natbib}
\usepackage[utf8]{inputenc} 
\usepackage[T1]{fontenc}    
\usepackage{hyperref}       
\usepackage{url}            
\usepackage{booktabs}       
\usepackage{amsmath} 
\usepackage{amsfonts}       
\usepackage{nicefrac}       
\usepackage{microtype}      
\usepackage[ruled,vlined]{algorithm2e}
\usepackage{graphicx}
\usepackage{caption}
\usepackage{subcaption}
\usepackage{authblk}

\title{Goal directed molecule generation using Monte Carlo Tree Search}

\author[1]{Anand A. Rajasekar}
\author[1,3,4]{Karthik Raman}
\author[2,3,4]{Balaraman Ravindran}

\affil[1]{Department of Biotechnology, IIT Madras}
\affil[2]{Department of Computer Science and Engineering, IIT Madras}
\affil[3]{Initiative for Biological Systems Engineering, IIT Madras}
\affil[4]{Robert Bosch Centre for Data Science and Artificial Intelligence (RBC-DSAI), 
IIT Madras}
\date{}                     
\begin{document}

\maketitle

\begin{abstract}
One challenging and essential task in biochemistry is the generation of novel molecules with desired properties. Novel molecule generation remains a challenge since the molecule space is difficult to navigate through, and the generated molecules should obey the rules of chemical valency. Through this work, we propose a novel method, which we call \textit{\textbf{unitMCTS}}, to perform molecule generation by making a unit change to the molecule at every step using Monte Carlo Tree Search. We show that this method outperforms the recently published techniques on benchmark molecular optimization tasks such as QED and penalized logP. We also demonstrate the usefulness of this method in improving molecule properties while being similar to the starting molecule. Given that there is no learning involved, our method finds desired molecules within a shorter amount of time. 
\end{abstract}

\section{Introduction}
One of the fundamental problems in chemistry is the design of novel molecules with specific properties. Molecule generation is essential for drug design and material science. Nevertheless, the task remains a difficult one due to the vastness of the chemical space. It is estimated that the number of drug-like molecules is around $10^{23}$ -- $10^{60}$ \citep{drugnum}. Also, the chemical space is discrete with high sensitivity of properties to molecular structure. Hence, there is an increasing need for efficient models focussed on specific applications that guide molecule generation. 

\noindent In recent years, various ML approaches have been deployed in molecular design. Generally, molecules are represented as SMILES strings \citep{acs, MI} (a linear string notation), fed as input to the model. One of the primary challenges with the SMILES notation is to ensure the validity of generated strings since the generated strings may or may not correspond to a molecule. A better way of representing molecules is with molecular
graphs where a node represents an atom, and an edge represents a bond. This method's advantage is the stepwise generation of molecular graphs \citep{li1,li2}, which is valid at every step. One common strategy for molecular graph generation is using a generative model such as Generative Adversarial Network or Variational Auto Encoder. Although generative models seem a viable option for generation, these models cannot optimize for a particular property. Hence, an additional step is usually employed that involves optimization in the latent space of the model using a property-oriented decoder \citep{nevae}, Particle Swarm Optimization \citep{mso}, or Bayesian Optimization \citep{jt}. Non-convexity and high dimensionality of the latent space usually render the task of optimization highly difficult. A second strategy is based on reinforcement learning, which involves an agent that interacts with the environment to learn and make decisions that maximize the cumulative reward. Again, the featurization of molecules here can be done with SMILES representation. Using RL techniques with SMILES representation \citep{organ, dnr} struggle in generating valid molecules. However, search-based techniques such as MCTS with SMILES can help maintain chemical validity \citep{mcts} by removing invalid SMILES strings. Using molecular graphs representation with RL provides 100\% validity \citep{gcpn}. All the methods mentioned above involves the use of a dataset. MolDQN \citep{moldqn} pursued a different approach by constructing a Markov Decision Process with unit modification to a molecule as actions and used DQN to approximate the molecules' value. Molecules are represented using fingerprints, and this method learns to generate molecules without the help of a dataset. PGFS \cite{cspace} proposed multiple modifications to an existing molecule at every step. This method constitutes an MDP with valid reactions as possible actions, while reactants and products are the states. We follow the MDP structure defined by MolDQN \citep{moldqn}; however, we do not allow bridge atoms so that the generated molecules look feasible. Algorithmically, our method combines Monte Carlo Tree Search, a well-established technique, and below mentioned formulation of MDP in a novel way, which is critical for its success. 

\section{Markov Decision Process}
We define the generation process for our method using the MDP, M = $\big(S, A, R, P \big)$ which is defined as follows. \\\\
\textbf{State space, $S$} consists of all valid molecules, $s$ $\in$ $S$. At $t=0$, $s_0$ begins with nothing or a specific molecule (for constrained optimization), and at every step, the molecule is modified. The number of steps allowed from the starting state is limited according to the task at hand. \\\\
\textbf{Action space, $A$} consists of all possible modifications that can be performed on a molecule. Given any molecule, all modifications that can be applied to it falls under one of four categories; 
    \begin{itemize}
        \item Atom addition - Let $K$ be the set of atoms that are allowed for addition. For any given molecule $M$, all atoms in set $K$ are added to molecule $M$ at every location that results in a chemically valid bond. These additions result in several possible new molecules.
        \item Bond addition - Any two atoms $a$ and $b$ that allow for a valid addition of bond in molecule $M$ is included in this list of molecules
        \item Bond removal - Any two atoms $a$ and $b$ connected by a bond in Molecule $M$ can be removed and included in this list of molecules. 
        \item Bond replacement - Any two atoms $a$ and $b$ connected by a bond can be replaced with another bond subject to valency conditions. All valid replacements are included in this list of molecules.
    \end{itemize}
Atom removal is not included here since when an atom becomes disconnected from a molecule, it is automatically removed. Together all these modifications correspond to actions that can be performed on the existing molecule $M$. We do not allow modifications that introduce bridge atoms in the molecule. \\\\
\textbf{Reward, $R$} associated with the generated molecule is used as feedback. We do not reward the system at every step. After each episode, properties associated with the final generated molecule are used to update the nodes' value. \\\\
\textbf{Transition probability, $P$} - Since the MDP is deterministic, each action can correspond to a single new state. Hence, the probability of reaching that state is one while all other states are zero.

\section{Proposed method}
In a regular tree search, one evaluates a node, its descendants, and so on until a final solution is obtained. However, this brute force search is not efficient for tasks that exponentially increase the number of nodes as we go further down the tree. One such example that has a vast number of possibilities is the game of Go. However, humans have achieved a state of the art performance in Go with the help of MCTS \cite{alphago}. MCTS makes the tree search faster by arriving at a policy that gives more importance to favorable descendants while lesser importance to others. This way, only a minimal number of nodes are explored in order to obtain the optimal solution. \\

\textbf{Monte Carlo Tree Search} consists of four steps, which are repeated for a certain number of times. The four steps are:
\begin{itemize}
    \item Selection - This step involves selecting a node that favors both exploration and exploitation among all descendants of a parent. The strategy used to select the node is called tree policy and is defined in Equation~\eqref{eq:1}.  
    \item Expansion - Once we reach the tree's leaf node, possible future states of the node are added to the tree, thus expanding it. 
    \item Simulation - After a new node is added, a simulation is performed for a specified rollout depth using a simple rollout policy.
    \item Backpropagation - The resulting state's value is then used to update $V$ and $n$ of nodes in the selected path of the tree. 
\end{itemize}
\begin{equation}
   child =  \underset{i}{argmax} \;\;\frac{V_i}{n_i} + c \sqrt{\frac{ln N_i}{n_i}} \label{eq:1}
\end{equation}
where $V_i$ is the value of the node, $n_i$ and $N_i$ are the numbers of times the node, and its parent have been visited, and $c$ is a hyperparameter that decides the importance of exploration and exploitation.

\begin{algorithm}[ht]
\SetAlgoLined
    \For {each mcts step}{
        \While{not leaf}{
            \textbf{Select} - Pick a child node based on tree policy in Equation~\eqref{eq:1} \\
        }
    \textbf{Expand} - Add top $k$ children to the tree and initialize the number of visits, $n$ with one and value, $V$ with reward obtained by $\epsilon$-greedy policy for each new node. Backpropagate value and visits till the root of the tree.\\
    \textbf{Simulate} - Execute the rollout policy until rollout depth and pick the final molecule.\\ 
    \While{not root}{
            \textbf{Backpropagate} - Scale the reward exponentially using a scaling factor $\alpha$. Increment $V$ and $n$ of the node in the selected path by the scaled reward of molecule and one, respectively. Move to the parent of the node.
    }
}
\caption{unitMCTS}
\label{alg:algo}
\end{algorithm}


\section{Results \& Discussion}

\textbf{Setup} - RDKit \citep{rdkit} is used for the molecule environment. The maximum steps per episode are 38 and 84 for \textbf{unitMCTS-38} and \textbf{unitMCTS-84} respectively, for all molecular optimization tasks starting from scratch and 20 for constrained optimization tasks, being set similar to MolDQN\citep{moldqn} for a fair comparison. \textit{\textbf{unitMCTS}} uses three atoms, i.e., Carbon, Nitrogen, and Oxygen for molecule generation. We optimize for two metrics, namely Quantitative Estimate of Drug Likeness (QED) and Penalized logP. QED is a weighted sum of a molecule's fundamental properties, such as its solubility, molecular weight, etc. It has a bounded range of [0,1]. Penalized logP is the logarithm of the partition ratio of solute between octanol and water subtracted by synthetic accessibility score and long cycles. It has a range of $(-\infty, \infty)$.

\textbf{Baselines} - We compare our method with the following baselines. JTVAE \citep{jt} uses a graph representation of molecules combined with Variational Auto Encoder for generating molecular graphs and Bayesian Optimization on latent space for optimizing property scores.  ORGAN \citep{organ} follows a text representation of molecules coupled with RL based generation. ChemTS \citep{mcts} uses the Monte Carlo Tree Search for SMILES generation with RNN based rollout policy. We also compare our work with GCPN \citep{gcpn}, which uses a graph convolutional network for molecule generation, and MolDQN \citep{moldqn}, which uses Bootstrapped DQN to estimate the value of molecules. Above mentioned baselines limit the number of steps per episode to 38. We also compare our method with PGFS \citep{cspace}, which proposes a forward synthesis framework powered by reinforcement learning, and MSO \citep{mso}, which generates molecules using Particle Swarm Optimization. To perform a fair comparison with the last two baselines, we set the number of steps per episode to 84, equal to the maximum bond number present in the dataset used by PGFS.\\\\
To measure the effectiveness of molecule generation using MCTS, we have evaluated our approach on two standard tasks, namely Property optimization and Constrained optimization. 
\subsection{Property optimization}
\begin{figure}
\centering
\begin{subfigure}{0.5\textwidth}
  \centering
  \includegraphics[width=0.7\linewidth]{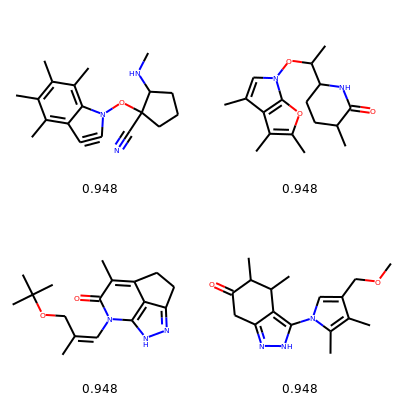}
  \caption{QED optimization}
  \label{fig:sub1}
\end{subfigure}%
\begin{subfigure}{0.5\textwidth}
  \centering
  \includegraphics[width=0.7\linewidth]{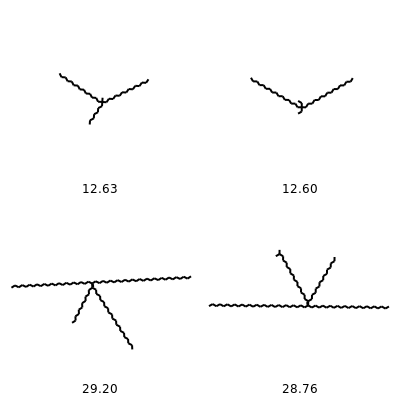}
  \caption{Penalized logP optimization}
  \label{fig:sub2}
\end{subfigure}
\caption{Top molecules generated by our method}
\label{fig:mol}
\end{figure}

This task aims to generate molecules with high QED and Penalized logP scores. Table~\ref{tab:prop} summarizes the property optimization results obtained by our method compared to other approaches. unitMCTS-38 outperforms all the baselines with an average improvement of 60\% over GCPN and 6\% over MolDQN-bootstrap on the Penalized logP task. It outperforms the baselines on the QED task as well. unitMCTS-84 outperforms both the baselines on the Penalized logP task. All of our generated molecules for both QED and Penalized logP tasks look realistic and are displayed in Figure~\ref{fig:mol}. Given that only valid molecules are added to the tree, our method's validity is 100\%. 
\begin{table}
    \caption{Top 3 molecules obtained by each method on QED and Penalized logP tasks}
    \label{tab:prop}
        \centering
    \begin{tabular}{l|llllll}
        \toprule
        & \multicolumn{3}{c}{Penalized logP} & \multicolumn{3}{c}{QED}\\
        \cmidrule(lr){2-7}
        {} & 1st & 2nd & 3rd  & 1st & 2nd & 3rd  \\
        \midrule
        ORGAN            & 3.63  & 3.49	 & 3.44	 & 0.896 & 0.824 & 0.82 \\
        JTVAE            & 5.3	 & 4.93  & 4.49  & 0.925 &	0.911 &	0.91\\
        ChemTS           & 6.56	 & 6.43  & 6.34  & --- & --- & ---\\
        GCPN             & 7.98	 & 7.85  & 7.8   & \textbf{0.948} & 0.947 & 0.946\\
        MolDQN-bootstrap & 11.84 & 11.84 & 11.82 &\textbf{0.948}  & 0.944 & 0.943\\
        Ours (unitMCTS-38) & \textbf{12.63} & \textbf{12.6} & \textbf{12.55} & \textbf{0.948} &
        \textbf{0.948} & \textbf{0.948}\\
        \midrule
        MSO             & 26.10	 & ---  & ---   & \textbf{0.948} & --- & ---\\
        PGFS & 27.22 & --- & --- &\textbf{0.948}  & --- & ---\\
        Ours (unitMCTS-84) & \textbf{29.20} & \textbf{28.76} & \textbf{28.73} & \textbf{0.948} &
        \textbf{0.948} & \textbf{0.948}\\

        \bottomrule
    \end{tabular}
\end{table}

\subsection{Constrained optimization}
In this task, we chose 800 molecules with the lowest Penalized logP scores in the ZINC \citep{zinc} dataset and performed a search to improve the Penalized logP while maintaining similarity with the starting molecule at various thresholds. Table~\ref{tab:pen} summarizes the performance of our method compared to the baselines. Our method outperforms the baselines at all thresholds. The tree expansion for this task is slightly different from the above tasks since only the top $k$ molecules that pass the similarity check are added to the tree. Hence, the success rate of our method is 100\%.



\begin{table}
    \caption{Mean and S.D. of Penalized logP improvement in constrained optimization tasks}
    \label{tab:pen}
        \centering
    \begin{tabular}{l|llllllll}
        \toprule
        $\delta$ & \multicolumn{1}{c}{JT-VAE} & \multicolumn{2}{c}{GCPN} & \multicolumn{2}{c}{MolDQN-bootstrap} & \multicolumn{2}{c}{Ours (unitMCTS-20)}\\
        \midrule

0.0	 & 1.91 $\pm$ 2.04 && 4.20 $\pm$ 1.28 && 7.04 $\pm$ 1.42 && 9.47 $\pm$ 2.38\\
0.2	 & 1.68 $\pm$ 1.85 && 4.12 $\pm$ 1.19 && 5.06 $\pm$ 1.79 && 8.21 $\pm$ 1.99\\
0.4	 & 0.84 $\pm$ 1.45 && 2.49 $\pm$ 1.30 && 3.37 $\pm$ 1.62 && 6.21 $\pm$ 1.15\\
0.6	 & 0.21 $\pm$ 0.71 && 0.79 $\pm$ 0.63 && 1.86 $\pm$ 1.21 && 3.44 $\pm$ 1.80\\
        \bottomrule
    \end{tabular}
\end{table}

\section{Future work}
Our work primarily deals with unit modification to a molecule at every step. Molecule modification with non-unit changes is one promising direction for future work where molecules are added instead of atoms. Our future work in this area is to propose MolMCTS that adds molecules at every step, similar to MDP proposed by PGFS \citep{cspace}.

\bibliography{neurips_2020}

\end{document}